\documentclass[runningheads]{llncs}
\usepackage[T1]{fontenc}
\usepackage{graphicx,verbatim}
\usepackage{multirow}
\usepackage[normalem]{ulem}
\useunder{\uline}{\ul}{}
\usepackage{adjustbox}
\usepackage{bbm}

\usepackage{graphicx}
\usepackage{amsmath}
\usepackage{amssymb}
\usepackage{mathtools}
\usepackage{booktabs}
\usepackage{times}
\usepackage{microtype}
\usepackage{epsfig}
\usepackage[table,xcdraw,dvipsnames]{xcolor}
\usepackage{caption}
\usepackage{float}
\usepackage{placeins}
\usepackage{color, colortbl}
\usepackage{stfloats}
\usepackage{enumitem}
\usepackage{tabularx}
\usepackage{xstring}
\usepackage{multirow}
\usepackage{xspace}
\usepackage{url}
\usepackage{subcaption}
\usepackage{xcolor}
\usepackage[hang,flushmargin]{footmisc}
\usepackage{hyperref}

\begin{document}
\title{PathoSCOPE: Few-Shot Pathology Detection via Self-Supervised Contrastive Learning and Pathology-Informed Synthetic Embeddings}
\titlerunning{PathoSCOPE}
%

\author{
  Sinchee Chin\inst{1} \and
  Yinuo Ma\inst{1} \and
  Xiaochen Yang\inst{2} \and
  Jing-Hao Xue\inst{3} \and
  Wenming Yang\inst{1}
}

\authorrunning{Chin et al.} 

\institute{
  \inst{1} Tsinghua Shenzhen International Graduate School \\
  \email{chenxz22@mails.tsinghua.edu.cn}, 
  \email{ma-yn24@mails.tsinghua.edu.cn}, 
  \email{yang.wenming@sz.tsinghua.edu.cn}
  \and
  \inst{2} University of Glasgow \\
  \email{xiaochen.yang@glasgow.ac.uk}
  \and
  \inst{3} University College London \\
  \email{jinghao.xue@ucl.ac.uk}
}

\maketitle              
\begin{abstract}
    Unsupervised pathology detection trains models on non-pathological data to flag deviations as pathologies, offering strong generalizability for identifying novel diseases and avoiding costly annotations. However, building reliable normality models requires vast healthy datasets, as hospitals’ data is inherently biased toward symptomatic populations, while privacy regulations hinder the assembly of representative healthy cohorts.
    To address this limitation, we propose \textbf{PathoSCOPE}, a few-shot unsupervised pathology detection framework that requires only a small set of non-pathological samples (minimum 2 shots), significantly improving data efficiency. 
    We introduce \textbf{Global-Local Contrastive Loss~(GLCL)}, comprised of a Local Contrastive Loss to reduce the variability of non-pathological embeddings and a Global Contrastive Loss to enhance the discrimination of pathological regions.
    We also propose a \textbf{Pathology-informed Embedding Generation (PiEG)} module that synthesizes pathological embeddings guided by the global loss, better exploiting the limited non-pathological samples.
    Evaluated on the BraTS2020 and ChestXray8 datasets, PathoSCOPE achieves state-of-the-art performance among unsupervised methods while maintaining computational efficiency (2.48 GFLOPs, 166 FPS). 
    The code will be made publicly available soon.

    
\end{abstract}

\section{Introduction}
\label{sec:intro}




Pathology detection in medical imaging refers to the task of identifying diseases in medical scans, such as MRIs and CT scans. Traditional approaches for pathology detection rely heavily on classification algorithms, which often require large amounts of labeled data, especially when dealing with a wide variety of pathologies~\cite{wang2024pathology}. However, obtaining annotated medical data is a major hurdle, particularly for rare diseases and under-represented conditions~\cite{colman2024real,de2024epidemiological}. 
Therefore, a more practical approach is needed. Inspired by industrial anomaly detection~(AD) methods~\cite{bao2024bmad,TMI-PATHOLOGY-DETECTION}, \textbf{Unsupervised Pathology Detection (UPD)} has recently gained significant traction in medical imaging. It shifts the focus from labeling a vast array of pathological conditions to modeling healthy samples and identifying anomalies as potential diseases, mitigating the need for extensive pathology labeling while maintaining clinical relevance. 

UPD can generally be classified into four approaches~\cite{TMI-PATHOLOGY-DETECTION}.
\textbf{Reconstruction-based methods} learn to reconstruct non-pathological samples and compute anomaly scores via perceptual loss between input and reconstructed healthy images~\cite{akcay2019ganomaly,schlegl2019f}. While effective, these methods often require a large number of training samples to avoid overfitting~\cite{xia2022gan,han2021madgan,esmaeili2023generative,zhao2023omnial}.
\textbf{Attention-based methods} enhance reconstruction-based methods by improving model's attention to normality through GradCAM~\cite{liu2020towards}. However, the use of GradCAM is suboptimal in few-shot scenario as it faces challenges of generalization, feature reuse, and class ambiguity, resulting in less precise heatmaps.
\textbf{Feature modeling methods} leverage Imagenet pre-trained neural networks as feature extractors to extract high-level feature representations for anomaly detection. However, these networks are pre-trained on natural images, which limits their effectiveness due to the domain disparity with medical images~\cite{defard2021padim,gudovskiy2022cflow}.
More recently, \textbf{self-supervised methods}, such as DRAEM~\cite{zavrtanik2021draem}, employ generative self-supervised learning to train a discriminator to recognize anomalies from self-generated anomalous samples.
SimpleNet~\cite{liu2023simplenet} and GLASS~\cite{chen2024unified} added a linear projection to transform embeddings into a learned space to address the domain disparity faced in feature modeling techniques and then trained a discriminator to recognize these synthetic anomalous embeddings. 
However, they encounter a critical bottleneck in generating synthetic data that fails to resemble reality. 
Although current UPD is effective, their performance declines with the \textbf{lack of non-pathological training samples} and \textbf{high inherent diversity of human anatomical structures}.

In this paper, we propose PathoSCOPE, a novel few-shot \textbf{Patho}logy detection method leveraging \textbf{S}elf-Supervised \textbf{Co}ntrastive learning and \textbf{P}athology-informed synthetic \textbf{E}mbeddings to addresses the challenge of detecting pathologies in a few shots and anatomically diverse medical images.
First, a pre-trained feature extractor is applied to images from healthy subjects to generate non-pathological anatomical features, forming the prototypical anchor bank. This ensures stability against the high variability of synthetic data. Next, a feature adapter is trained to regularize synthetic non-pathological embeddings, referring to the prototypical features while enhancing discriminability to synthetic pathological features through Global-Local Contrastive Learning~(GLCL). This dual objective enhances the separation between normal and pathological regions by regularizing variations in non-pathological features and amplifying deviations indicative of pathology.
To synthesize anatomically plausible pathological features, we propose a Pathology-informed Embedding Generation~(PiEG) module that applies targeted perturbations to the non-pathological embeddings. These perturbations mimic real-world pathological features while preserving contextual coherence with surrounding anatomy. Finally, a discriminator is trained to identify pathological regions by detecting deviations from normal features.
In sum, our contributions in this paper are threefold:
\begin{itemize}
    \item We propose \textbf{Global-Local Contrastive Loss~(GLCL)} to regularize the high variance of non-pathological embeddings and guide synthetic pathological embeddings generation.
    \item We propose \textbf{Pathology-informed Embedding Generation~(PiEG)} module, effectively generate pathological embeddings that mimic real-world pathology patterns.
    \item We achieve state-of-the-art performance in few-shot UPD on BraTS2020 and ChestXray8 datasets while maintaining high throughput and low GFLOPs, paving the way for clinical deployment.
\end{itemize}

\section{Method}
\label{sec:method}

\subsection{Overview}
\label{sec:method-overall}
\begin{figure*}[tp]
    \centering
    \includegraphics[width=0.8\linewidth]{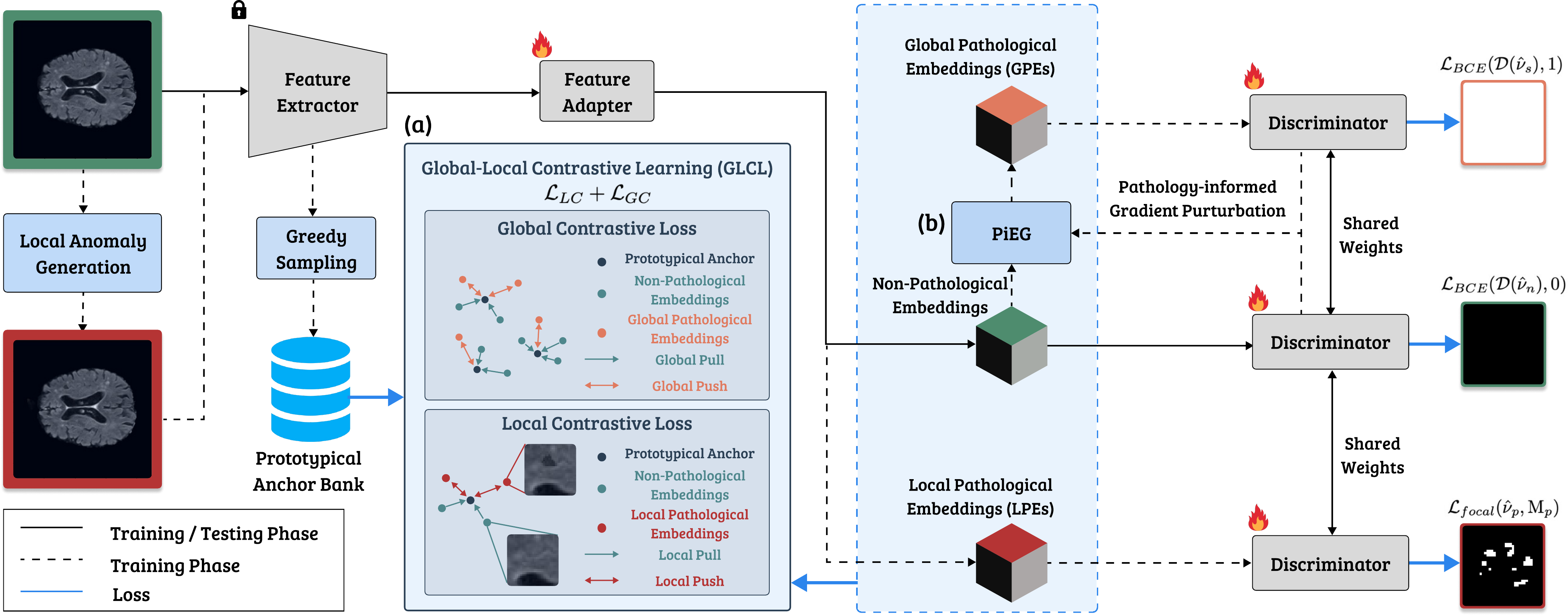}
    \caption{Overall architecture of the PathoSCOPE}
    \label{fig:overall}
\end{figure*}

The overview of PathoSCOPE is shown in Fig.~\ref{fig:overall}.
PathoSCOPE consists of three parts: a pre-trained feature extractor, $\mathbf{E}_\phi:\mathbb{R}^{H \times W \times C} \rightarrow \mathbb{R}^{L_H \times L_W \times L_C}$, used to extract features from the input image, a feature adapter, $\mathbf{f}_\phi:\mathbb{R}^{L_H \times L_W \times L_C} \rightarrow \mathbb{R}^{L_H \times L_W \times L_S}$, used to map the extracted embeddings to a learned space, and a discriminator, $\mathbf{D}_\phi:\mathbb{R}^{L_H \times L_W \times L_S} \rightarrow \mathbb{R}^{L_H \times L_W \times 1}$, used to provide a pathology score for each pixel of the learned embeddings. 

PathoSCOPE introduces two simple yet significant improvements: \textbf{(a) Global-Local Contrastive Learning (GLCL)} anchors embeddings to a Prototypical Anchor Bank of non-pathological features, reducing anatomical variance by enforcing consistency between local regions; and, \textbf{(b) Pathology-informed Embedding Generation (PiEG) module} synthesizes global pathological embeddings (GPEs) by perturbing non-pathological embeddings using the discriminator’s gradients. This closes the loop between anomaly synthesis and detection, ensuring GPEs mimic real pathologies.

During inference, PathoSCOPE predicts the pathological regions by passing the input image through the feature extractor, feature adapter, and discriminator, achieving high FPS and low GFLOPs.
The obtained pathological map is then interpolated to the input image size, $(H,W)$.
Lastly, the image is classified as pathological if the maximum anomaly score in the pathological map exceeds a predefined threshold, $\tau$.

\subsection{Feature Extractor and Feature Aggregation}
\label{sec:method-contrastive}
Prior to training, we leverage the feature extractor, specifically, a ResNet18 backbone, to extract image embeddings from non-pathological samples, $x_n$. The embeddings are obtained by concatenating features from the second and third layers. 
Using the coreset sampling algorithm~\cite{sener2017active}, we select the prototypical embeddings and store them in an external prototypical anchor bank, $\mathbf{\Phi}_A$, which serves as stable anatomical references during model training.

During training, we perform data augmentation on the training samples with random affine transformations and elastic deformations to generate additional non-pathological images, $\hat{x}_n \in \mathbb{R}^{H \times W \times C}$, where $H$, $W$, $C$ is the height, width, and channel of the image. These augmentations simulate the internal body movement and symmetry of the human anatomy~\cite{zhao2019data,garcea2023data}. Then, $\hat{x}_n$ is passed to the local anomaly generation module to generate local pathological images, $\hat{x}_p$. We use the DTD texture dataset~\cite{cimpoi2014describing} to ensure diversity and randomness in the synthetic pathological features. The local anomaly generation module also outputs a local pathological map, $\mathrm{M}_p \in \{0, 1\}$, with $1$ labeled as region of simulated pathology and vice versa.

The augmented non-pathological images are passed to the feature extractor and then to the trainable feature adapter to project the non-pathological embeddings to a learned embedding space: $\hat{\nu}_n=\mathbf{f}_\phi(\mathbf{E}_\phi(\hat{x}_n))$. Using the same procedure, we obtain Local Pathological Embeddings (LPEs), $\hat{\nu}_p$. We then compute the distances of $\hat{\nu}_n$ and $\hat{\nu}_p$ with the closest prototypical anchor point from $\mathbf{\Phi}_A$ using Eq. \ref{eq:local-push-pull}:
\begin{equation}
    \label{eq:local-push-pull}
    d_{\text{local}}(\hat{\nu}) = \frac{1}{\sum_{h=1}^{L_h}\sum_{w=1}^{L_w}\mathrm{M}_p(h,w)} \sum_{h=1}^{L_h}\sum_{w=1}^{L_w} \mathrm{M}_p(h,w) \cdot\min{\lVert \hat{\nu}{(h,w)} - \mathbf{\Phi}_{A} \rVert_2},
\end{equation}

Next, we leverage the Tritanh loss, Eq. \ref{eq:tritanh-loss}, for Global-Local Contrastive Loss: 
\begin{equation}
\label{eq:tritanh-loss}
    \mathcal{L}_{C}(d_{\text{pull}}, d_{\text{push}}) = \frac{
    e^{\lambda_0 \cdot d_{\text{pull}}} - 
    e^{\lambda_1 \cdot d_{\text{push}}} + 
    \epsilon
    }{
    e^{\lambda_0 \cdot d_{\text{pull}}} + 
    e^{\lambda_1 \cdot d_{\text{push}}} + 
    \epsilon
    },
\end{equation}
where $\epsilon$ is a regularizing term and $\lambda_0$ and $\lambda_1$ are the scaling factors for the pulling and pushing force respectively.
The larger the scaling factor, the more emphasis is placed on that particular feature in the contrastive learning process. The Tritanh loss provides a more robust gradient flow, bounded loss function, and unique solution, mitigating catastrophic collapse and gradient explosion \cite{reiss2023mean,chen2024unified}.

The Local Contrastive Loss is computed with Eq.~\ref{eq:tritanh-loss} as follows: $\mathcal{L}_{LC}=\mathcal{L}_C(d_{\text{local}}(\hat{\nu}_n),d_{\text{local}}(\hat{\nu}_p))$, which trains the feature adapter to \textbf{regularize high variance embeddings} by minimizing $d_{\text{local}}(\hat{\nu}_n)$ and \textbf{enhance subtle discriminability} by maximizing $d_{\text{local}}(\hat{\nu}_p)$.

\subsection{Pathology-informed Embedding Generation}
\label{sec:method-ctgn}
\begin{figure}[tp]
    \centering
    \begin{subfigure}[t]{0.48\textwidth}
        \centering
        \includegraphics[width=0.7\linewidth]{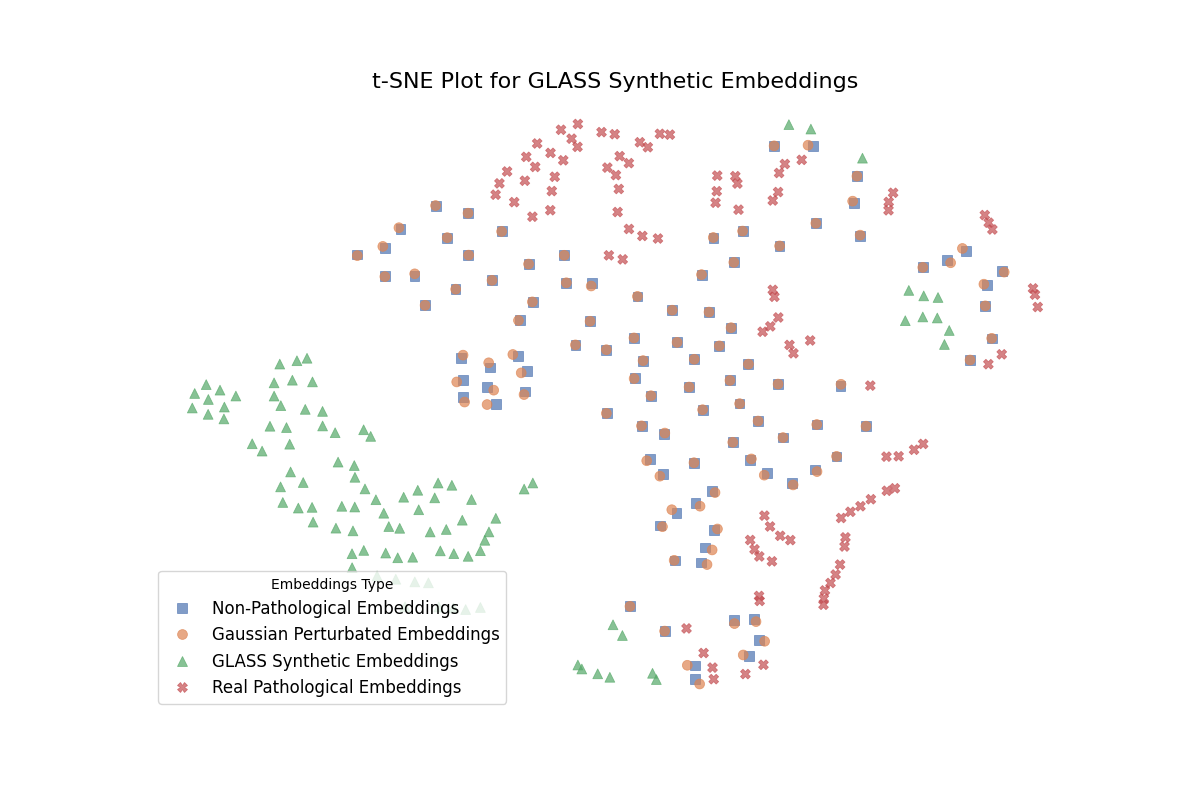} 
        \caption{GLASS-h}
        \label{fig:gradient-noise-glass}
    \end{subfigure}
    \hfill 
    \begin{subfigure}[t]{0.48\textwidth}
        \centering
        \includegraphics[width=0.7\linewidth]{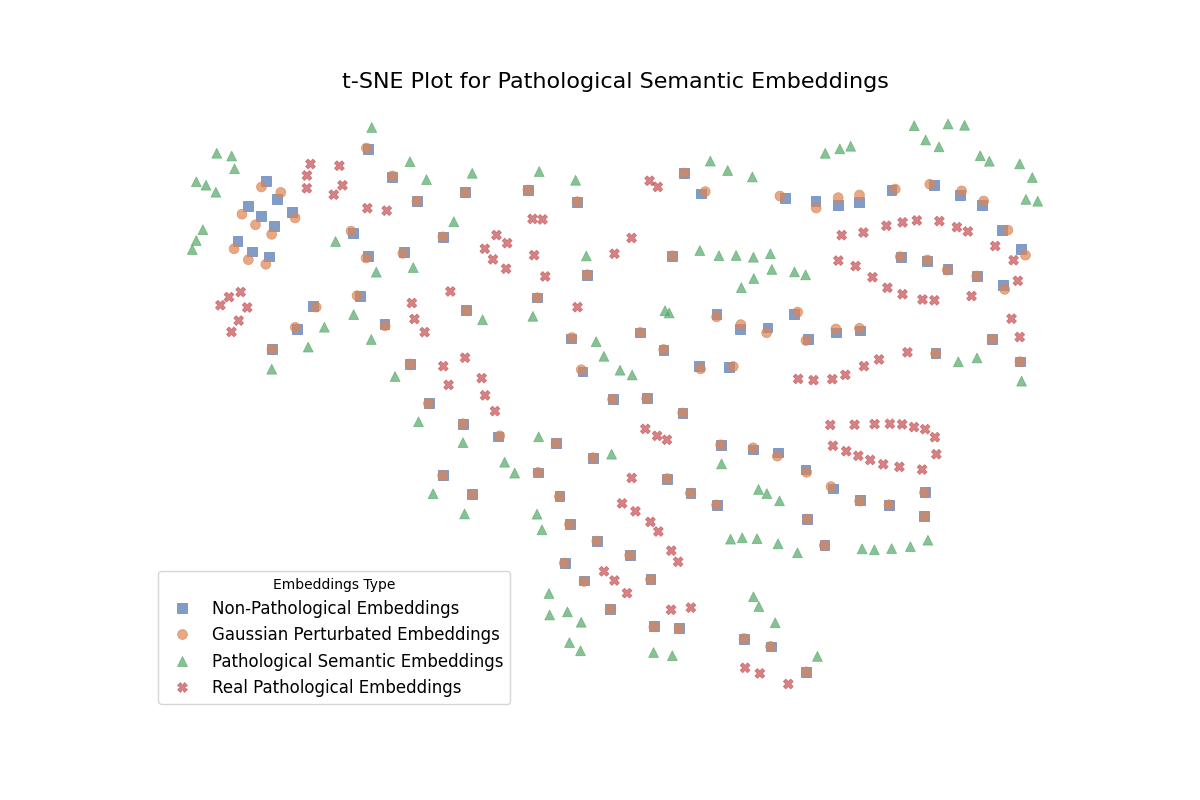} 
        \caption{PathoSCOPE}
        \label{fig:gradient-noise-glam}
    \end{subfigure}
    \caption{Comparison of t-SNE for synthetic embeddings on ChestXray8 dataset. The pathology-informed embeddings generated by PathoSCOPE more closely resemble real pathological embeddings than GLASS synthetic embeddings.}
    \label{fig:tsne-mtgn}
\end{figure}
While $\mathcal{L}_{LC}$ focuses on enhancing local discrimination and feature regularization, it does not account for the varying difficulty of pathological embeddings. To generate harder, more realistic embeddings, we propose the Pathology-informed Embedding Generation (PiEG) module to synthesize Global Pathological Embeddings (GPEs), which will then be trained through Global Contrastive Loss. This design enables holistic embedding regularization without the need for real pathological data. 

We initialize the GPEs by adding random Gaussian noises to non-pathological embeddings: $\mathbf{\hat{\nu}}_{(s,0)}=\mathbf{\hat{\nu}}_n+\rho$, where $\rho \sim \mathcal{N}(\mu, \Sigma) \in \mathbb{R}^{L_H \times L_W \times L_C}$. The GPEs are then iteratively updated over $T$ steps, accumulating gradients from the Global Loss, $\mathcal{L}_{\text{global}}$ (will be defined in Eq. \ref{eq:global-loss}), as described in Eq.~\ref{eq:metric-based-gaussian-noise}: 
\begin{equation}
    \label{eq:metric-based-gaussian-noise}
    \hat{\nu}_s \coloneq \mathbf{\hat{\nu}}_{(s,T)} = \mathbf{\hat{\nu}}_n + \eta\sum_{i=0}^{T-1} \frac{\nabla \mathcal{L}_{\text{global}}(\mathbf{\hat{\nu}}_n, \mathbf{\hat{\nu}}_{(s,i)})}{\lVert \nabla \mathcal{L}_{\text{global}}(\mathbf{\hat{\nu}}_n, \mathbf{\hat{\nu}}_{(s,i)}) \rVert},
\end{equation}
where $\eta$ denotes the perturbation strength, controlling the magnitude of deviation applied to non-pathological embeddings, $\hat{\nu}_n$ when synthesizing GPEs. Practically, we set $T=20$, $\mu=0.015$, and $\Sigma=I$. Fig.~\ref{fig:tsne-mtgn} shows the t-SNE plot comparing synthetic embeddings generated by GLASS~\cite{chen2024unified} and PathoSCOPE. 

Next, we compute the Global Contrastive Loss by first calculating the distances of $\hat{\nu}_s$ and $\hat{\nu}_n$ to the closest prototypical anchor point from $\mathbf{\Phi}_A$ using Eq.~\ref{eq:global-push-pull} and then applying the Tritanh loss in Eq.~\ref{eq:tritanh-loss} with $\mathcal{L}_{GC} = \mathcal{L}_C(d_{\text{global}}(\hat{\nu}_n), d_{\text{global}}(\hat{\nu}_s))$: 
\begin{equation}
    \label{eq:global-push-pull}
    d_{\text{global}}(\hat{\nu}) = \frac{1}{L_h \times L_w} \sum_{h=1}^{L_h}\sum_{w=1}^{L_w} \min{\lVert \hat{\nu}{(h,w)} - \mathbf{\Phi}_{A} \rVert_2}.
\end{equation}
As Eq.~\ref{eq:global-push-pull} does not include the local pathological map $\mathrm{M}_p$, the Global Contrastive Loss focuses on optimizing the distances at the semantic level.

\subsection{Discriminator and Loss Function}

To train the discriminator and feature adapter jointly, we compute the Local Loss and Global Loss. The Local Loss, $\mathcal{L}_{\text{local}}$, combines the focal loss and Local Contrastive Loss, as shown in Eq.~\ref{eq:local-loss}:
\begin{equation}
\label{eq:local-loss}
    \mathcal{L}_{\text{local}} = \mathcal{L}_{focal}(\hat{\nu}_p, \mathrm{M}_p) + \mathcal{L}_{LC}
\end{equation}

The Global Loss, $\mathcal{L}_{\text{global}}$, combines the binary cross-entropy (BCE) loss, which explicitly separates non-pathological features from PGEs, and the Global Contrastive Loss as shown in Eq.~\ref{eq:global-loss}:

\begin{equation}
    \label{eq:global-loss}
    \begin{split}
        \mathcal{L}_{\text{global}}(\hat{\nu}_n,\hat{\nu}_s) = &\; \mathcal{L}_{BCE}(\mathcal{D}(\hat{\nu}_n), 0) + \mathcal{L}_{BCE}(\mathcal{D}(\hat{\nu}_s), 1) + \mathcal{L}_{GC}
    \end{split}
\end{equation}

The overall loss function is the sum of $\mathcal{L}_\text{global}$ and $\mathcal{L}_\text{local}$.

\section{Results and Discussion}
\label{sec:results}
\subsection{Datasets}
We benchmark our method using two medical datasets: BraTS2020~\cite{menze2014multimodal,bakas2017advancing,bakas2018identifying} and the ChestXray8~\cite{wang2017chestx}. To simulate a few-shot scenario, we randomly select 2 to 8 healthy subjects for training and ensure no overlap between the training and test datasets.
For the BraTS2020 dataset, MRI slices are randomly sampled from slice 60 to slice 90, resulting in 130 healthy samples and 290 tumorous samples in the test set; only one MRI slice is sampled for each training subject.
For the ChestXray8 dataset, we randomly select 2 to 8 subjects for training and evaluate performance on 1,490 test samples.
All experiments are repeated 10 times, and the average metrics are reported in Tables \ref{tab:main-brats2020} and \ref{tab:result-chest}.
\begin{table*}[tp]
\centering

\caption{
Performance comparison of PathoSCOPE on the BraTS2020 dataset for different number of shots (2, 4, 6, and 8 shots). All experiments are repeated 10 times.
}

\resizebox{\textwidth}{!}{%

\begin{tabular}{cc|c|c|cccc|cccc|cccc|cccc}
\hline
\multicolumn{2}{c|}{\multirow{2}{*}{Method}}      & \multirow{2}{*}{FPS (img / s)} & \multirow{2}{*}{GFLOPs} & \multicolumn{4}{c|}{Image AUROC (\%)}                             & \multicolumn{4}{c|}{Pixel AUROC (\%)}                             & \multicolumn{4}{c|}{PRO (\%)}                                     & \multicolumn{4}{c}{DICE (\%)}                                     \\
\multicolumn{2}{c|}{}                             &                                       &                         & K=2            & K=4            & K=6            & K=8            & K=2            & K=4            & K=6            & K=8            & K=2            & K=4            & K=6            & K=8            & K=2            & K=4            & K=6            & K=8            \\ \hline
\multirow{2}{*}{Image Reconstruction} & f-AnoGAN ~\cite{schlegl2019f}   & 27                                    & 5.78              & 59.29          & 57.73          & 57.87          & 54.49          & -              & -              & -              & -              & -              & -              & -              & -              & -              & -              & -              & -              \\
                                      & Ganomaly ~\cite{akcay2019ganomaly}  & 3                                     & 16.51              & 46.20          & 43.29          & 42.98          & 45.72          & -              & -              & -              & -              & -              & -              & -              & -              & -              & -              & -              & -              \\ \hline
\multirow{5}{*}{Feature Modeling}    & PaDIM ~\cite{defard2021padim}    & 18                                    & 45.18                   & 66.07          & 72.90          & 73.58          & 75.13          & 92.28          & 94.50          & 94.55          & 94.89          & 65.78          & 73.87          & 74.00          & 74.70          & 27.67          & 36.57          & 36.86          & 38.30          \\
                                      & CFLOW ~\cite{gudovskiy2022cflow}     & 64                                    & 703.98                  & 80.91          & 82.42          & 82.99          & 84.52          & 94.95          & 94.79          & 94.68          & 94.73          & 77.85          & 76.67          & 76.07          & 74.20          & 42.71          & 39.63          & 39.30          & 40.94          \\
                                      & RD ~\cite{deng2022anomaly}        & 78                                    & 831.74                  & 44.61          & 47.19          & 46.86          & 47.36          & 15.23          & 15.40          & 15.76          & 15.75          & 0.00           & 0.00           & 0.00           & 0.00           & 5.36           & 5.38           & 5.37           & 5.38           \\
                                      & RegAD ~\cite{huang2022registration}    & 95                                    & 189.73                  & 64.39          & 62.08          & 70.04          & 68.35          & {\ul 95.95}    & {\ul 95.83}    & {\ul 95.97}    & {\ul 96.17}    & {\ul 86.34}    & {\ul 84.05}    & {\ul 84.85}    & {\ul 85.28}    & 34.51          & 34.85          & 34.00          & 34.81          \\
                                      & FAE ~\cite{meissen2022unsupervised}      & 16                                     & 6.67              & 60.20          & 70.51          & 68.40          & 69.85          & 91.09          & 93.24          & 93.04          & 93.35          & 72.53          & 77.54          & 77.29          & 77.86          & 26.11          & 32.02          & 31.43          & 33.26          \\ \hline
\multirow{2}{*}{Attention-Based}      & ExpVAE ~\cite{liu2020towards}   & 24                            & 3.07              & 57.75          & 55.46          & 59.72          & 60.15          & 67.92          & 47.51          & 54.13          & 55.72          & 36.49          & 15.77          & 21.73          & 26.03          & 23.98          & 10.13          & 17.77          & 13.17          \\
                                      & AMCons ~\cite{silva2022constrained}   & 24                            & 123.89              & 56.80          & 54.33          & 58.79          & 53.53          & 93.09          & 91.47          & 93.48          & 93.39          & 73.89          & 72.04          & 75.15          & 74.42          & 43.78          & {\ul 41.84}    & {\ul 46.04}    & {\ul 44.52}    \\ \hline
\multirow{5}{*}{Self-Supervised}      & DRAEM ~\cite{zavrtanik2021draem}    & 96                                    & 4867.06                 & 60.36          & 59.79          & 56.87          & 57.55          & 94.44          & 95.01          & 95.07          & 94.93          & 75.29          & 74.75          & 75.38          & 75.02          & 39.62          & 38.52          & 40.72          & 38.81          \\
                                      & SimpleNet ~\cite{liu2023simplenet} & {\ul 163}                            & {\ul 3.06}              & 62.27          & 62.27          & 65.03          & 65.07          & 64.54          & 58.53          & 57.75          & 50.80          & 41.30          & 41.20          & 40.80          & 41.20          & 13.60          & 13.60          & 13.40          & 13.50          \\
                                      & GLASS-h \cite{chen2024unified}  & \textbf{166}                         & \textbf{2.48}           & {\ul 87.39}    & {\ul 88.42}    & {\ul 88.66}    & {\ul 87.19}    &   95.63    & 95.17          & 94.57          & 92.68          & 80.51          & 78.79          & 78.58          & 76.45          & 41.16          & 38.29          & 36.94          & 35.02          \\
                                      & GLASS-m \cite{chen2024unified}  & \textbf{166}                         & \textbf{2.48}           & 86.66    & 85.88          & 87.05      & 87.00    & 94.12          & 93.24          & 94.20          & 92.72          & 84.82          & 79.47          & 80.14          & 82.94          & {\ul 46.31}    & 38.93          & 37.20          & 41.57          \\
                                      & \textbf{Ours}       & \textbf{166}                         & \textbf{2.48}           & \textbf{89.19} & \textbf{89.96} & \textbf{89.01} & \textbf{87.84} & \textbf{97.87} & \textbf{97.79} & \textbf{97.55} & \textbf{97.38} & \textbf{88.95} & \textbf{89.06} & \textbf{89.06} & \textbf{87.75} & \textbf{49.21} & \textbf{48.02} & \textbf{48.02} & \textbf{44.63} \\ \hline
\end{tabular}

}

\label{tab:main-brats2020}
\end{table*}

\begin{table*}[tp]
\centering
\begin{minipage}{0.48\textwidth} 
\centering
\caption{Performance comparison of PathoSCOPE on the ChestXray8 dataset for different number of shots (2, 4, 6, and 8 shots). Image AUROC (\%) is reported.}
\label{tab:result-chest}
\resizebox{0.8\linewidth}{!}{%
\begin{tabular}{cc|cccc}
\hline
\multicolumn{2}{c|}{\multirow{2}{*}{Method}}      & \multicolumn{4}{c}{Image AUROC (\%)}                              \\
\multicolumn{2}{c|}{}                             & K=2            & K=4            & K=6            & K=8            \\ \hline
\multirow{2}{*}{Image Reconstruction} & f-AnoGAN  & 41.63          & 40.36          & 39.52          & 41.45          \\
                                      & Ganomaly  & 60.55          & 55.43          & 57.43          & 55.70          \\ \hline
\multirow{5}{*}{Feature Modelling}    & PaDIM     & 32.65          & 31.93          & 31.25          & 30.70          \\
                                      & CFLOW     & 31.22          & 30.97          & 29.98          & 28.95          \\
                                      & RD        & 68.26          & 66.98          & 65.37          & 63.57          \\
                                      & RegAD     & 66.06          & 64.06          & 65.26          & 64.24          \\
                                      & FAE       & 64.95          & 67.94          & 71.72          & 70.84          \\ \hline
\multirow{2}{*}{Attention Based}      & ExpVAE    & 55.35          & 55.68          & 54.56          & 55.16          \\
                                      & AMCons    & 42.15          & 42.43          & 42.92          & 41.95          \\ \hline
\multirow{5}{*}{Self Supervised}      & DRAEM     & 56.38          & 57.43          & 51.74          & 54.98          \\
                                      & SimpleNet & 65.82          & 67.70          & 68.22          & 67.07          \\
                                      & GLASS-h   & 69.52          & {\ul 69.53}    & {\ul 72.44}    & {\ul 70.49}    \\
                                      & GLASS-m   & {\ul 70.03}    & 69.49          & 70.27          & 68.50          \\
                                      & Ours      & \textbf{72.23} & \textbf{73.64} & \textbf{73.94} & \textbf{72.95} \\ \hline
\end{tabular}%
}
\vfill 
\end{minipage}%
\hfill
\begin{minipage}{0.48\textwidth}
\centering
\caption{Ablation study of GLCL on BraTS2020. Reported metrics are Image AUROC / Pixel AUROC.}
\label{tab:ablation-contrastive-brats2020}
\resizebox{\textwidth}{!}{%
\begin{tabular}{c|c|c|c|c}
\hline
\multirow{2}{*}{Contrastive Loss} & \multicolumn{4}{c}{Number of Shot} \\ \cline{2-5}
                                  & K=2               & K=4               & K=6               & K=8               \\ \hline
None                              & 86.66 / 94.12    & 85.88 / 93.24    & 87.05 / 94.20    & 87.00 / 92.72    \\
$\mathcal{L}_{LC}$                             & 84.11 / 95.53    & 87.57 / 96.22    & 86.49 / 95.11    & 86.04 / 94.15    \\
$\mathcal{L}_{GC}$                            & 84.09 / 95.53    & 87.57 / 96.22    & 80.88 / 91.34    & 86.04 / 94.15    \\ \hline
$\mathcal{L}_{LC} + \mathcal{L}_{GC}$                    & \textbf{89.19 / 97.87} & \textbf{89.96 / 97.79} & \textbf{89.01 / 97.55} & \textbf{87.84 / 97.38} \\ \hline
\end{tabular}%
}

\caption{Ablation study of GLCL on ChestXray8. Image AUROC is reported.}
\label{tab:ablation-contrastive-chestxray}
\resizebox{0.6\linewidth}{!}{%
\begin{tabular}{c|c|c|c|c}
\hline
\multirow{2}{*}{Contrastive Loss} & \multicolumn{4}{c}{Number of Shot} \\ \cline{2-5}
                                  & K=2               & K=4               & K=6               & K=8               \\ \hline
None                              & 70.03 &	69.49 &	70.27 &	68.50    \\
$\mathcal{L}_{LC}$                             & 71.73 &	71.74 &	{\ul 72.13} &	71.29    \\
$\mathcal{L}_{GC}$                            & {\ul 71.76} &	{\ul 71.77} &	72.06 &	{\ul 71.33}    \\ \hline
$\mathcal{L}_{LC} + \mathcal{L}_{GC}$                    & \textbf{72.23} & \textbf{73.64} & \textbf{73.94} & \textbf{72.95} \\ \hline
\end{tabular}%
}
\end{minipage}
\end{table*}


\subsection{Evaluation Metrics} 
We evaluate PathoSCOPE with four metrics. 
\textbf{Image AUROC} measures the ability to classify images as normal or pathological.
\textbf{Pixel AUROC} evaluates anomaly localization at the pixel level.
\textbf{Pixel Per Region Overlap (PRO)} balances detection across regions, ensuring smaller regions are not overshadowed by larger ones \cite{yang2020improving}. 
\textbf{DICE Score} measures the similarity between the predicted anomaly mask and the ground truth mask. 
We also compute the \textbf{Intersection Over Union (IoU)} and show it on top of the visualization.

\subsection{Discussion}
PathoSCOPE consistently achieves the best performance across different shots on both BraTS2020 dataset and ChestXray8 datasets, especially in segmentation task. 
PathoSCOPE also has the lowest GFLOPs and the high FPS compared to previous methods. 
 
\begin{figure}[t]
    \centering
    \includegraphics[width=0.5\linewidth]{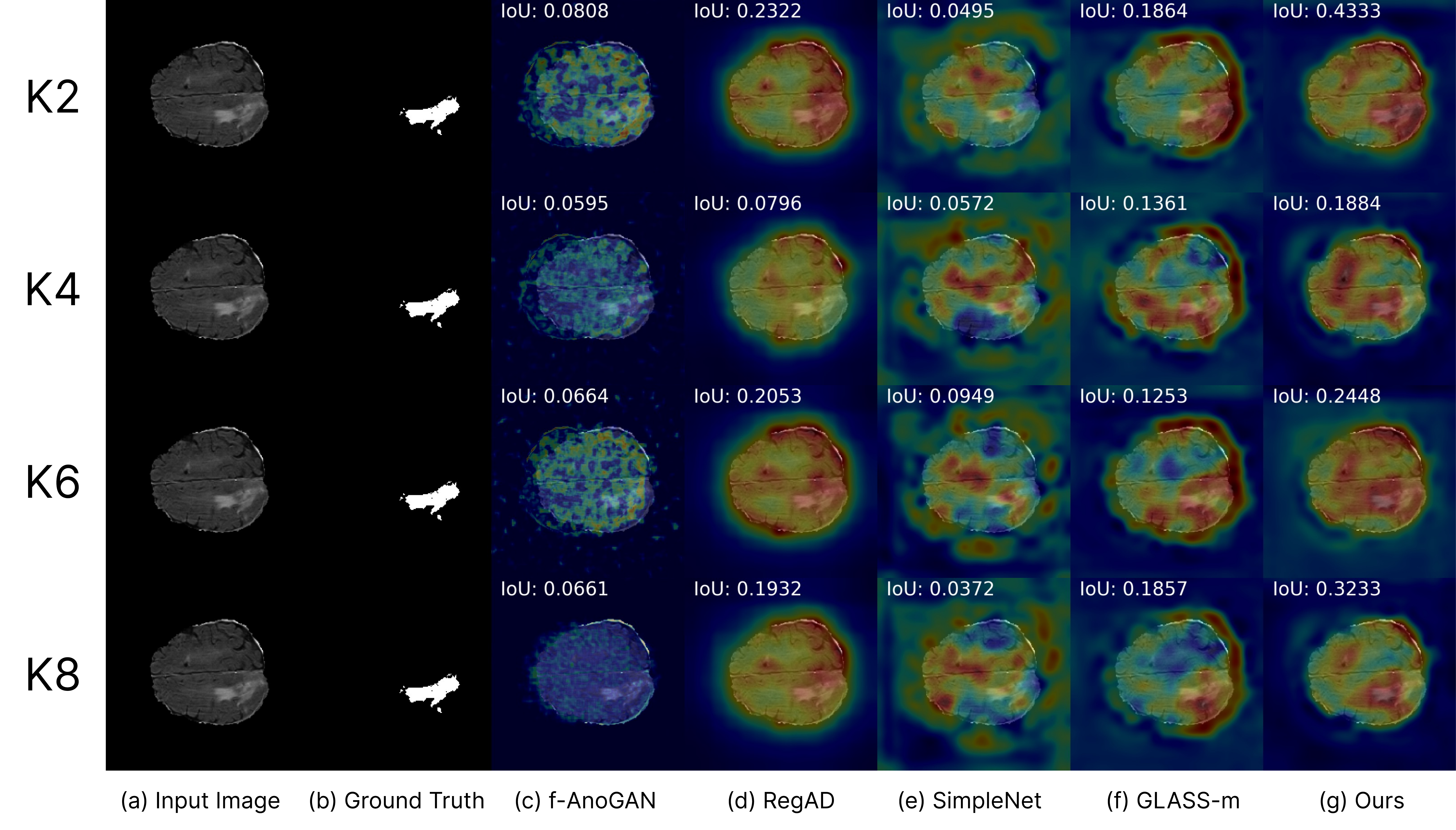}
    \caption{Visualization of BraTS2020 Heatmap Prediction}
    \label{fig:dicussion-visualize}
\end{figure}

\begin{figure*}[tp]
    
    \centering

    \begin{subfigure}{0.32\textwidth}
        \centering
        \includegraphics[width=0.9\linewidth]{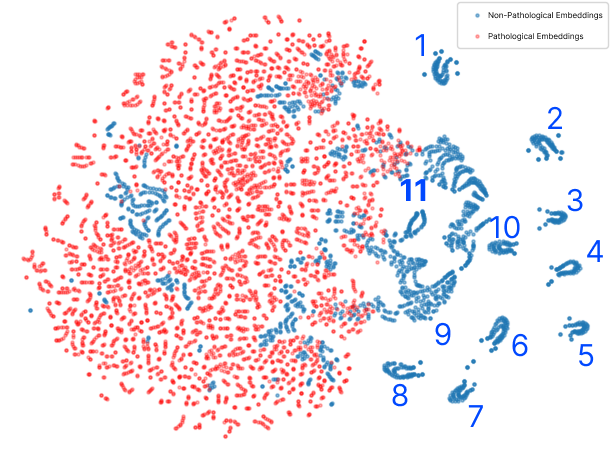}
        \caption{GLASS-h (11 Clusters)}
        \label{fig:tsne-glass-h}
    \end{subfigure}
    \hfill
    \begin{subfigure}{0.32\textwidth}
        \centering
        \includegraphics[width=0.9\linewidth]{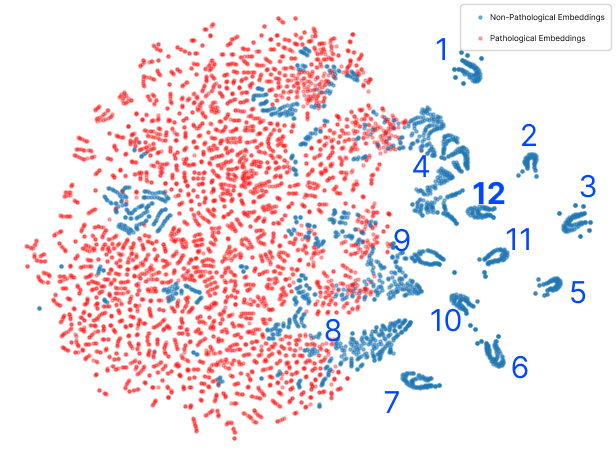}
        \caption{GLASS-m (12 Clusters)}
        \label{fig:tsne-glass-m}
    \end{subfigure}
    \hfill
    \begin{subfigure}{0.32\textwidth}
        \centering
        \includegraphics[width=0.9\linewidth]{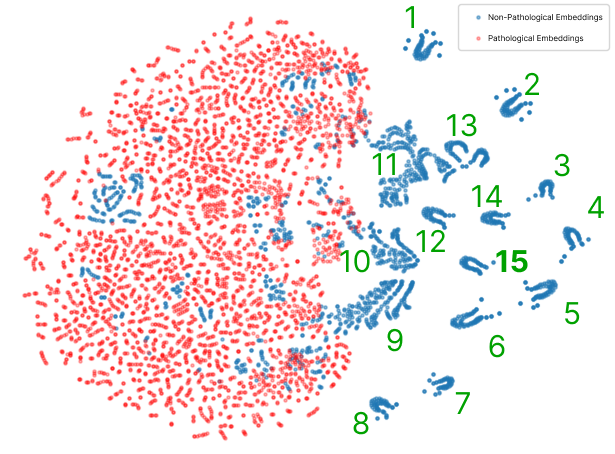}
        \caption{PathoSCOPE (15 Clusters)}
        \label{fig:tsne-pathoscope}
    \end{subfigure}

    \caption{Comparison of t-SNE for (a) GLASS-Hypersphere, (b) GLASS-Manifold, and (c) PathoSCOPE.}
    \label{fig:compare-tsne}
\end{figure*}

The visual results in Fig. \ref{fig:dicussion-visualize} highlight PathoSCOPE’s ability to detect subtle pathologies under varying few-shot settings. Unlike baseline methods, which produce fragmented or over-smoothed anomaly maps, PathoSCOPE maintains sharp boundaries and anatomical coherence. This precision is further validated by the t-SNE analysis of BraTS2020 test set in Fig. \ref{fig:compare-tsne}, where PathoSCOPE’s embeddings exhibit distinct non-pathological clusters and clearer separation from pathological regions. We attribute this to the GLCL, which suppresses non-pathological feature variance while amplifying discriminative signals, and the PiEG module, which generates synthetic pathologies that closely mimic real-world pathology distributions.

These results underscore PathoSCOPE’s effectiveness in addressing two key challenges: (1) the \textbf{inherent diversity of human anatomy}, mitigated through prototypical feature regularization, and (2) the \textbf{scarcity of pathological training data}, alleviated by PiEG’s anatomy-aware synthesis. The framework’s efficiency and accuracy position it as a practical solution for clinical deployment, particularly in resource-constrained settings requiring rapid, few-shot adaptation.

\subsection{Ablation Study}

\noindent\textbf{Global-Local Contrastive Loss:}
Further experiments on BraTS2020 and ChestXray8 (Tables~\ref{tab:ablation-contrastive-brats2020} and~\ref{tab:ablation-contrastive-chestxray}) demonstrate the necessity of combining $\mathcal{L}_{LC}$ and $\mathcal{L}_{GC}$.
While $\mathcal{L}_{LC}$ reduces variance in non-pathological embeddings by contrasting local regions, it fails to propagate gradients to \textit{all} non-pathological features during optimization. This results in inconsistent regularization, leaving some non-pathological features under-constrained.
The adversarial feedback is critical to refining GPEs not only from the $\mathcal{L}_{GC}$ but also through iterative updates guided by $\mathcal{L}_{LC}$ derived from prior iterations. Without $\mathcal{L}_{LC}$, GPEs fail to capture the spatially coherent features of healthy features, generating overrepresented anomalies.
The joint optimization of $\mathcal{L}_{LC}$ and $\mathcal{L}_{GC}$ is crucial to holistic non-pathological feature regularization and realistic pathological embedding generation.



\begin{figure}[tbp]
    \centering
    \begin{subfigure}[b]{0.48\textwidth}
        \centering
        \includegraphics[width=0.6\textwidth]{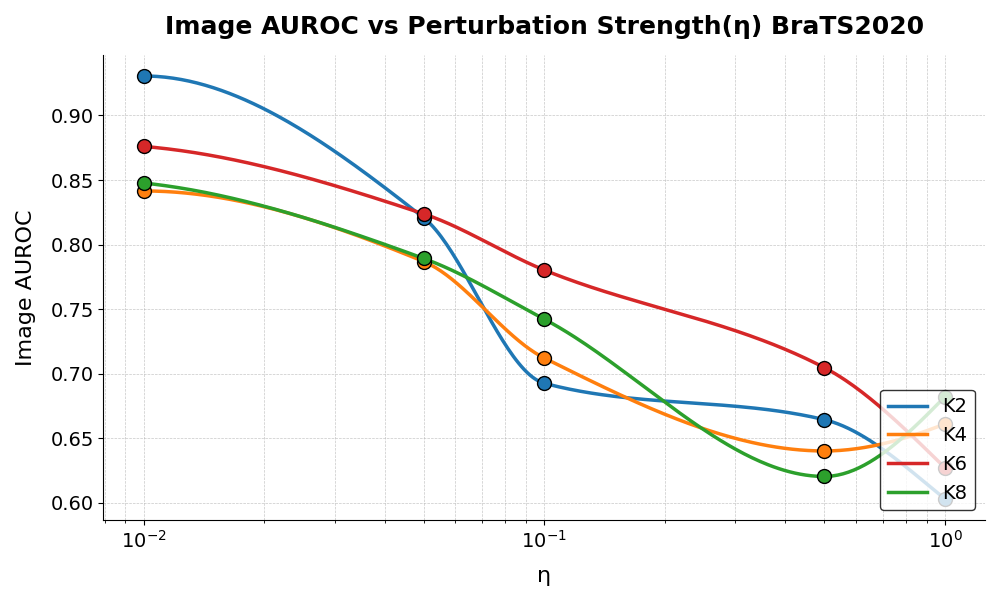}
        \caption{BraTS2020}
        \label{fig:subfig1}
    \end{subfigure}
    \begin{subfigure}[b]{0.48\textwidth}
        \centering
        \includegraphics[width=0.6\textwidth]{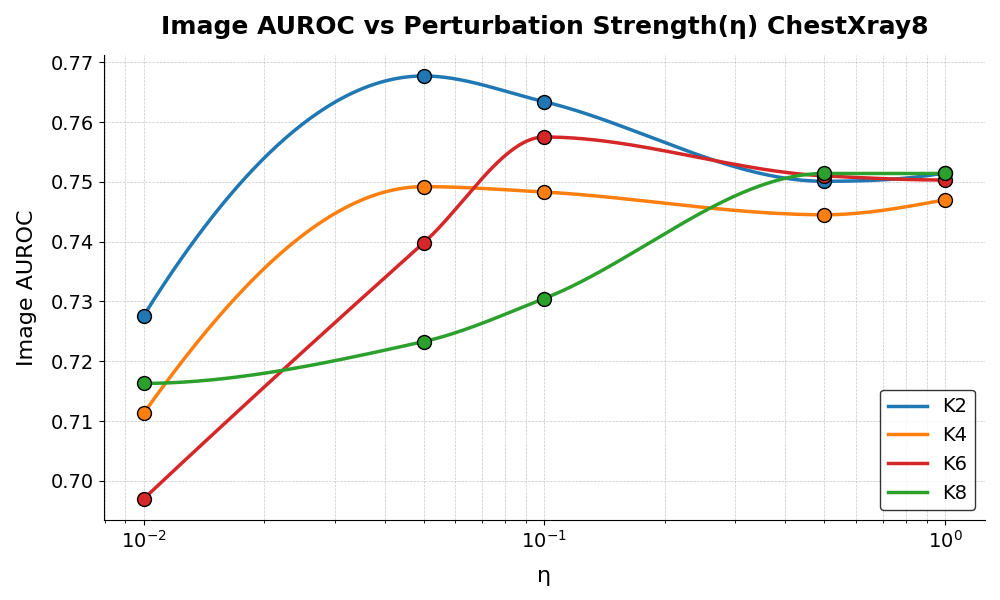}
        \caption{ChestXray8}
        \label{fig:subfig2}
    \end{subfigure}
    \caption{Evaluation of the perturbation strength ($\eta$) on BraTS2020 and ChestXrat8 datasets.}
    \label{fig:ablation-eta}
\end{figure}

\noindent\textbf{Perturbation strength, $\eta$, of PiEG:} 
As $\eta$ increases, the GPEs drift further from the non-pathological distribution, simulating more severe anomalies (Fig. \ref{fig:ablation-eta}). On BraTS2020 (brain MRI), $\eta=0.01$ achieves peak performance by perturbing embeddings just enough to mimic subtle lesions while preserving anatomical plausibility. For ChestXray8, a higher perturbation strength leads to better performance ($\eta=0.05$), likely due to the coarser texture and larger anomaly regions (e.g., consolidations) in chest radiographs. Exceeding these values degrades AUROC, as excessive distortion creates implausible features that mislead the discriminator. This mirrors a key challenge in medical anomaly synthesis: pathologies must "deviate" from normal anatomy without violating its inherent structural coherence.

\section{Conclusion}
\label{sec:conclusion}
PathoSCOPE effectively addresses the lack of non-pathological training samples and anatomical diversity in current UPD, achieving accurate anomaly detection with as few as two non-pathological training samples. PathoSCOPE’s computational efficiency (2.48 GFLOPs, 166 FPS) further underscores its suitability for real-world clinical deployment.
One limitation of PathoSCOPE is that the generation of $\hat{x}_p$ lacks explicit anatomical coherence thus limiting their resemblance to real localized pathologies.
Future directions include training a decoder to synthesize $\hat{x}_p$ directly from the discriminator’s gradients via adversarial training, eliminating reliance on auxillary datasets, and improving anatomical realism.
Overall, PathoSCOPE represents a promising step toward efficient, few-shot UPD in medical imaging, with substantial potential for broader clinical applications such as novel disease recognition.

\bibliographystyle{splncs04}
\bibliography{11_references.bib}

\end{document}